\pdfoutput=1

\documentclass[11pt]{article}

\usepackage[]{acl}

\usepackage{times}
\usepackage{latexsym}

\usepackage[T1]{fontenc}

\usepackage[utf8]{inputenc}

\usepackage{microtype}

\usepackage{inconsolata}

\usepackage{graphicx} 
\usepackage[utf8]{inputenc}
\usepackage{amsmath}
\usepackage{url}

\title{Share What You Already Know: Cross-Language-Script Transfer and Alignment for Sentiment Detection in Code-Mixed Data}

\author{Niraj Pahari \and Kazutaka Shimada \\
  Kyushu Institute of Technology \\
  680-4 Kawazu, Iizuka, Fukuoka 820-8502, Japan \\
  \texttt{pahari.niraj828@mail.kyutech.jp} \and \texttt{shimada@ai.kyutech.ac.jp} }

\begin{document}
\maketitle
\begin{abstract}
Code-switching entails mixing multiple languages.
It is an increasingly occurring phenomenon in social media texts.
Usually, code-mixed texts are written in a single script, even though the languages involved have different scripts.
Pre-trained multilingual models primarily utilize the data in the native script of the language.
In existing studies, the code-switched texts are utilized as they are.
However, using the native script for each language can generate better representations of the text owing to the pre-trained knowledge.
Therefore, a cross-language-script knowledge sharing architecture utilizing the cross attention and alignment of the representations of text in individual language scripts was proposed in this study.
Experimental results on two different datasets containing Nepali-English and Hindi-English code-switched texts, demonstrate the effectiveness of the proposed method.
The interpretation of the model using model explainability technique illustrates the sharing of language-specific knowledge between language-specific representations.
\end{abstract}

\section{Introduction}
\label{sec:introduction}
\label{sec:introduction}
Humans use different languages to communicate with each other.
Communication can take various forms, including oral, written, or gestural.
When it comes to written communication, each language typically has its script in which it is written and read.
Many languages share the same script.
Multilingualism enables individuals to understand and use multiple languages.
This facilitates the mixing of different linguistic elements of different languages.
Although formal settings have more rigid language norms, code-mixing has become prevalent in informal or casual settings.

This phenomenon is also increasingly observed in written communication.
Informal written communication, even when characterized by code-mixing tends to have a consistent script.
This implies that, often, individuals express ideas in one language but use the script of another.
Figure \ref{fig:code_mixing_example} shows text written in romanized script that involves the mixing of Nepali and English language.

\begin{figure}[!t]
    \centering
    \fbox{
    \parbox[c]{0.45 \textwidth}{
        \textbf{Nepali-English:} Arule this is bad bhaneni malai chai ramro lagyo\\
        \\
        \textbf{Translation:} Even though others said, ``This is bad.", I liked it.
    }
    }
    \caption{Code-mixing example: Nepali-English pair and its English translation.}
    \label{fig:code_mixing_example}
\end{figure}

The official script of Nepali and Hindi is Devanagari.
However, in social media settings, individuals often prefer to use the romanized script.
This preference for romanized script is mainly owing to the education system in Nepal where computer typing is primarily taught for English language \cite{maharjan-etal-2015-developing}.
Furthermore, the learning curve of typing Devanagari is larger, as it involves a larger number of characters (48 primary characters including vowels and consonants), compared to the Latin/Roman alphabet (26) for which regular keyboards are designed.

Most pre-trained language models are trained on data from datasets that mostly contain the data in the official script of the language such as Wikipedia.
Even when the models are pre-trained in another script, the ratio of this kind of data is comparatively very low \cite{khanuja2021muril}.
Therefore, these models generally under-perform when faced with code-mixed data \cite{pires-etal-2019-multilingual}.
This is particularly challenging in sentiment analysis tasks, which entails analyzing digital text to detect emotion from it.
In sentiment analysis, understanding the emotion conveyed by individual words is important for accurate predictions.
Hence, there is a need for a model that can understand the semantic meaning of text, even when it is expressed in another script.

To overcome this need, we propose a model that leverages the benefit of different scripts when code-mixing is involved.
In the case of multimodal application, the dual encoder model proposed by Yu et al. \cite{yu-etal-2022-dual} uses cross-modal alignment of image-text pairs to utilize the information encoded for different modals.
Drawing inspiration from this model, we propose a language-script specific encoder model with cross-language-script alignment of information.
This enabled us to leverage the unique information in the representation of each script.
We evaluated the performance of the proposed model using two datasets comprising code-mixed Nepali-English and Hindi-English data.
The findings illustrated the superiority of cross-language-script transfer and alignment, to conventional baseline systems.
Furthermore, our examination of model interpretation revealed insights into the sharing of script-specific learning for the downstream task.
The error analysis yielded valuable information about the model's limitations, emphasizing the necessity of an in-depth investigation into the cleaning of romanized data to enhance the overall performance.

\section{Related Works}

Various approaches, ranging from traditional machine learning to advanced deep learning techniques, have been explored for sentiment detection in code-switched text \cite{mikolov2013efficient, vilares2015sentiment, invento2017sentiment, laureano-de-leon-etal-2020-cs}.
De Leon et al. \cite{laureano-de-leon-etal-2020-cs} introduced code-switched embeddings utilizing the word2vec \cite{mikolov2013efficient} model with the CBoW algorithm, complemented by a BiLSTM for classification.
Similarly, Angel et al. \cite{angel-etal-2020-nlp} employed a convolutional neural network for sentiment analysis in code-switched text.
These models exhibited susceptibility to biases introduced by cue words, particularly when expressed in English.
Srinivasan \cite{srinivasan-2020-msr} fine-tuned transformer-based multilingual models, such as mBERT and XLM-R, using code-switched data, showcasing the effectiveness of these models in handling code-switched content.
In a different approach, Liu et al. \cite{liu-etal-2020-kk2018} utilized adversarial training with XLM-R, introducing perturbations to input embeddings to generate new adversarial sentences, combined with an ensemble of diverse candidate models.
Ghosh et al. \cite{ghosh2023multitasking} employed multitask-learning for sentiment and emotion detection tasks, leveraging shared knowledge to enhance task-specific performance.

However, it is noteworthy that these existing studies are primarily useful for text in romanized form, and most pre-trained models are trained on native scripts for each language.
Ahmed et al. \cite{ahmad2023elevating} tackled spelling variations, multiple languages, different scripts, limited resources and similar challenges, by leveraging the auditory information of words to pre-train existing multilingual models. 
This approach demonstrated improved performance and enhanced robustness against adversarial attacks.

In the context of Nepali, which is considered a low-resource language spoken by more than 20 million individuals \cite{niraula2022linguistic}, there has been limited research within the sentiment analysis domain \cite{gupta2015detecting, regmi2017analyzing, tamrakar2020aspect}.
The pioneer study on Nepali sentiment analysis was conducted by Gupta and Bal \cite{gupta2015detecting}, as acknowledged by Shahi and Sitaula \cite{shahi2022natural}.
The authors constructed a sentiment analysis dataset sourced from diverse news portals and presented preliminary results using bag-of-words representations.
Additionally, Thapa and Bal \cite{thapa2016classifying} created document-level sentiment analysis utilizing book and movie reviews.
Piryani et al. \cite{piryani2020sentiment} employed deep learning and traditional machine learning models to conduct sentiment analysis on Nepali language tweets.
The authors also released Nepali SentiWordNet and Nepali SenticNet sentiment lexicon, derived from pre-existing English language resources.
Furthermore, Singh et al. \cite{singh2020aspect} curated a dataset for targeted aspect-based sentiment analysis and provided baseline methods utilizing mBERT and BiLSTM models.
The dataset predominantly comprised text in Devanagari script, with a mixture of romanized Nepali or English words and Devanagari words in a ratio of 0.0185.
These prior researchers primarily focused on Devanagari script, whereas we focused on code-mixed data containing romanized Nepali and English text.
We leveraged romanized and Devanagari representations by cross-language-script knowledge transfer.

In the domain of cross-modal knowledge sharing, Yu et al. \cite{yu-etal-2022-dual} employed a cross-attention method, facilitating knowledge exchange between text and image, two distinct representation spaces.
AlignVD \cite{chen2022unsupervised} model aligns text(dialogs) and image utilizing unsupervised and pseudo-supervised vision-language alignment.
The model uses graph autoencoder and dialog-guided visual grounding for alignment.
Similarly, language alignment has been extensively researched \cite{dyer-etal-2013-simple, abulkhanov2023lapca, li2023multilingual}.
In these approaches, embeddings in BERT-based multilingual models are aligned at the language level \cite{cao2020multilingual, kulshreshtha-etal-2020-cross}. 
This alignment targets zero-shot settings, aligning embeddings from one language to another with the goal of transferring knowledge acquired during pre-training. 
Pires et al. \cite{pires-etal-2019-multilingual} explored the ability of BERT-based multilingual models to transfer pre-trained knowledge between languages written in different scripts, demonstrating successful cross-script transfer.
However, challenges arise when the transfer involves languages with different syntactic structures. 
For instance, Nepali and Hindi follow the subject, object, verb order, whereas English follows the subject, verb, object order. 
Alignment alone may not suffice for effective transfer, prompting our proposal to transliterate romanized forms into Devanagari and leverage the knowledge from both scripts. 
We employed a cross-attention method to facilitate knowledge exchange between romanized and Devanagari representations, enhancing this process through the alignment of representations in both scripts.

\section{Language-Script Specific Multi-Encoder Model}
The proposed model was designed to improve the end task performance of code-mixed tasks by providing specific pre-trained model for the script of each of the language in the code-mixed text.
The advantage of this model is that the input token for each language is preserved in its native script, consistent with the script on which the model was originally trained, alongside the script of the mixed language.
Hence, the semantic meaning captured during the pre-training in specific script can be utilized for the end task.


\begin{figure*}[t]
    \centering
    \includegraphics[width=\textwidth]{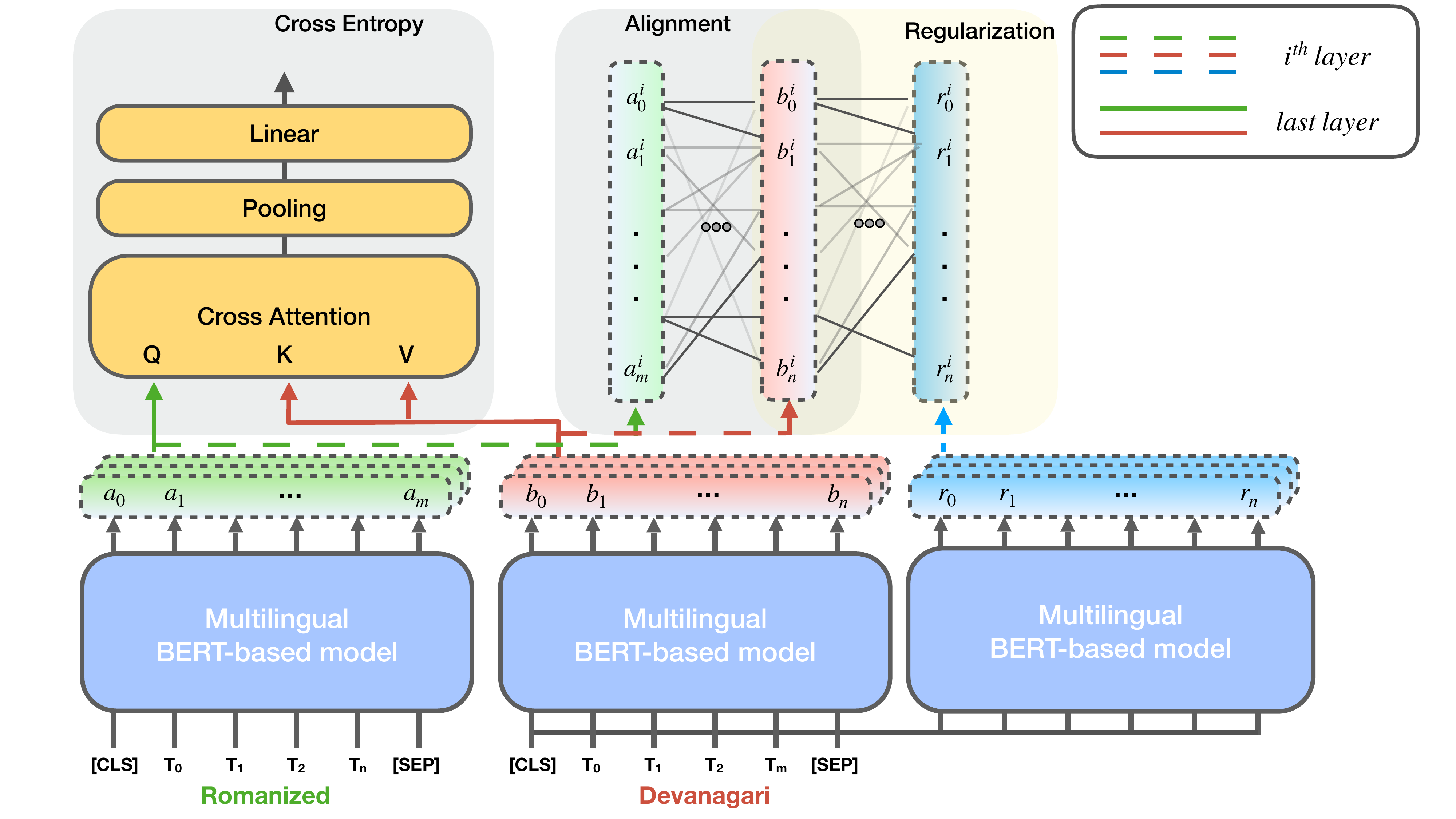}
    \caption{Proposed architecture of cross-language-script transfer multi-encoder model. The architecture comprises three modules: i. Cross-language-script transfer, as explained in Subsection \ref{sub:cross-language-script-transfer}, ii. Cross-script alignment, as explained in Subsection \ref{sub:cross-script-alignment}, and iii. Regularization, as explained in Subsection \ref{sub:regularization}.}
\label{fig:main_architecture}
\end{figure*}

Figure \ref{fig:main_architecture} shows the architecture of the proposed multi encoder model with script alignment.
The architecture comprised three transformer-based models.
Each individual transformer-based model was the same variant of the BERT-based multilingual pre-trained language model.
Two of these models were employed as the base for each language script and one extra model was used for the regularization of the alignment.
The architecture is explained in detail in subsequent sections.

\subsection{BERT-Based Multilingual Model}
BERT-based multilingual models were selected owing to their proven effectiveness across various multilingual NLP tasks \cite{pires-etal-2019-multilingual, wang2019cross}.
Multilingual Representations for Indian Languages (MuRIL) \cite{khanuja2021muril} was utilized in this study as the base pre-trained model.
The model was trained on English and 16 different languages spoken in and around India.
The model training was performed using monolingual, translated, and transliterated (to Latin/Roman) data.
The amount of translated and transliterated data used in pre-training was significantly less than the amount of monolingual data.
Masked language modeling objective was utilized for monolingual data, whereas, translation language modeling objective was utilized for both translated and transliterated document pairs.
The model was pre-trained on languages from the Indian subcontinent and the utilization of transliterated data during the pre-training phase, which makes it suitable for our experiments.

\subsection{Cross-Language-Script Transfer of Representations}
\label{sub:cross-language-script-transfer}
Each of the language script-specific base model facilitated representations based on the knowledge of the text in that script.
We applied cross-attention to these language script specific representations to share the semantic knowledge obtained during text pre-training in their correct script.
Let $H_A^l = \{a_{CLS}^l, \: a_0^l, \: a_1^l, \: ..., \: a_m^l\}$ be the last-layer representation of language $A$, and $H_B^l = \{b_{CLS}^l, \: b_0^l, \: b_1^l, \: ..., \: b_n^l\}$ be the last-layer representation of $B$.
Multi-head cross-attention was applied to share the knowledge between the representations of both language scripts.
The representation of Language $A$ is passed as query, whereas the representation of Language $B$ is passed as the key and the value.
The multi-head attention \cite{vaswani2017attention} is hence defined as

\begin{equation}
\begin{split}
&MultiHead(Q,\: K,\: V) \\ &= MultiHead(H_A^l,\: H_B^l,\: H_B^l)\\
    &= concat(head_1,\: ...,\: head_{N})W^O
	\end{split}
\end{equation}
where 
\begin{equation}
    \begin{split}
        & head_i = Attention(QW_i^Q,\: KW_i^K,\: VW_i^V)\\
        & = Attention(H_A^lW_i^Q,\: H_B^lW_i^K,\: H_B^lW_i^V)\\ \\
        & = Softmax\bigg(\frac{(H_A^lW_i^Q)(H_B^lW_i^K)^T}{\sqrt{d_h/N}}\bigg) (H_B^l W_i^V)
    \end{split}
\end{equation}

Here, $W_i^Q \in \operatorname{Re}^{d_h/N\times m}$, $ \{W_i^K, W_i^V\} \in  \operatorname{Re}^{d_h/N\times n}$,
$N$ is the number of attention heads, which was set to eight in this study, and $d_h$ is the hidden representations size which was $768$ for the model used in the study.
The output obtained from the multi-head attention layer was then forwarded to the pooling and linear layer.
Finally, the prediction $\hat{y}$ was obtained, and the cross entropy loss was calculated utilizing the true labels $y$ of the input.

\begin{equation}
    \begin{split}
    & \mathcal{L}_{CE}  = - \frac{1}{N} \sum_{i=1}^{N} \left[ y_i \cdot \log(\hat{y}_i) + (1 - y_i) \cdot \log(1 - \hat{y}_i) \right]
    \end{split}
\end{equation}

\subsection{Cross-Script Alignment}
\label{sub:cross-script-alignment}
Together with cross-attention between the language script-specific representations, we aimed to align the representation spaces between these representations.
Drawing inspiration from Zhang et al. \cite{zhang-etal-2017-earth}, our approach to cross-script alignment involved minimizing the earth mover's distance (EMD).
The choice of EMD was motivated by its capacity to comprehensively measure the closeness between two sets of weighted points without the need for parallel supervised data.

Intuitively, the EMD can be defined as the measure of the minimal effort required to transfer the earth mass distributed across a space to a set of designated holes within that space.
In our context, the earth mass and holes represent two distributions, and the EMD serves as a metric to measure the distance between these distributions.

The $i$-th layer hidden representations from both language specific models was utilized for alignment, where $i$ is an empirically-chosen hyperparameter.
Various studies \cite{jalili-sabet-etal-2020-simalign, dou-neubig-2021-word, pahari2022multitask, Pahari_2024jaciii} have explored the multiple layers of language models, revealing that the contextualization in the early layers tends to be insufficient for alignments, while the final layers become overly specialized for the end task.
Hence, we chose the use the $i$-th layer hidden representations instead of the last layer.

The $i$-th layer hidden representation of Language ${A}$ and Language ${B}$ with moving weight assigned can be represented as:
$P = \{(a_1^i, w_{a1}), ..., (a_m^i, w_{am})\}$ and $Q = \{(b_1^i, w_{b1}), ..., (b_n^i, w_{bn}\}$, respectively, where $w_{ax}$ and $w_{ay}$ were initialized as $1/m$ and $1/n$ respectively.
$D=[dxy]$ is the ground distance between the clusters $a_x$ and $b_y$, which we calculated using the mean squared error.
We aimed to find a flow $F=[f_{xy}]$, where $f_{xy}$ was the flow between $a_x$ and $b_y$, such that it minimizes the overall cost:
\begin{equation}
WORK(P, Q, F) = \sum_{x=1}^m\sum_{y=1}^nf_{xy}d_{xy}
\end{equation}
Subject to constraints
\begin{equation}
\label{eqn:emd-cons1}
    f_{xy} \ge 0, \qquad 1\le x \le m, \: 1\le y \le n  
\end{equation}
\begin{equation}
\label{eqn:emd-cons2}
    \sum_{y=1}^n f_{xy} \le w_{ax},   \qquad 1\le x \le m   
\end{equation}
\begin{equation}
\label{eqn:emd-cons3}
    \sum_{x=1}^m f_{xy} \le w_{by},   \qquad 1\le y \le n   
\end{equation}
\begin{equation}
\label{eqn:emd-cons4}
    \sum_{x=1}^m\sum_{y=1}^nf_{xy} = min(\sum_{x=1}^mw_{ax}, \sum_{y=1}^nw_{by})
\end{equation}

Where Constraint \ref{eqn:emd-cons1} allowed flow from $P$ to $Q$ and not vice versa.
Constraints \ref{eqn:emd-cons2} and \ref{eqn:emd-cons3} constrained the amount of features sent by $P$ and received by $Q$ did not exceed their weights.
Finally,Constraint \ref{eqn:emd-cons4} forced the maximum amount of features possible to flow.
Once we had found the optimal flow F, the EMD or the alignment loss was defined as the work normalized by the total flow:
\begin{equation}
\label{eqn:emd}
    \mathcal{L}_{SA} = EMD(P, Q) = \frac{\sum_{x=1}^m\sum_{y=1}^nf_{xy}d_{xy}}{\sum_{x=1}^m\sum_{y=1}^nf_{xy}}
\end{equation}

\subsection{Regularization}
\label{sub:regularization}
To address the alignment objective described above, the ideal solution is to equalize the representations of both language scripts.
However, equality will result in the loss of semantic information acquired during the pre-training of the language model.
To mitigate this, we introduced the regularization loss following \cite{cao2020multilingual, kulshreshtha-etal-2020-cross}.
We introduced a separate pre-trained model dedicated to regulating the representations of the Devanagari script, preventing them from deviating significantly.
This regularization was applied to the Devanagari script owing to its superior performance, compared to the romanized script in the baseline model of MuRIL.
The EMD of the $i$-th layer representation, as explained in Subsection \ref{sub:cross-script-alignment}, was employed for regularization.
Hence, the loss function was similar to \ref{eqn:emd}, as shown in \ref{eqn:reg}:
\begin{equation}
\label{eqn:reg}
    \mathcal{L}_{REG} = EMD(P, Q) = \frac{\sum_{x=1}^m\sum_{y=1}^nf_{xy}d_{xy}}{\sum_{x=1}^m\sum_{y=1}^nf_{xy}}
\end{equation}
Here, $P$ and $Q$ are the $i$-th layer representations of the Devanagari script and regularization model, $f_{xy}$ is the flow between $p_x$ and $q_y$, and $d_{xy}$ is the ground distance between $p_x$ and $q_y$.

\subsection{Final Training Objective}
The final training objective was the sum of the main task and the alignment task.
Specifically, the alignment task was a combination of script alignment and regularization subtasks.
The following equation represents our final objective:
\begin{equation}
Final\: Loss(\mathcal{L})= \mathcal{L}_{CE} \: +\alpha \: \times \: \mathcal{L}_{Align}    
\end{equation}
\begin{equation}
\mathcal{L}_{Align}= \beta\: \times\: \mathcal{L}_{SA} + \gamma \:\times \:\mathcal{L}_{REG}  
\end{equation}

where $\alpha$, $\beta$, and $\gamma$ were the hyperparameters that weight the contributions of each task.

\subsection{Model Interpretation}
To understand the decision-making of our proposed model, we employed an explainable AI technique known as SHapley Additive exPlanations (SHAP) \cite{lundberg2017unified}.
This method is based on the cooperative game theory which attracts interdisciplinary interest owing to its exploration of the fundamental role of cooperation in human society, nature, or model performances \cite{sharma2023small}.
It entails evaluating the contribution of each input feature for the final result by calculating Shapley values.

SHAP is premised on the consideration that each input feature makes a certain contribution to the model's prediction.
A perturbation-based approach is adopted, where variations in inputs are introduced, and the resulting changes in the output are analyzed as explanations.
Specifically, SHAP hides individual words from the input sentence, submits this modified sentence to the target model, and observes the corresponding alterations in the output probabilities.
This process is repeated for all possible combinations of word hiding and, finally, the relevance of each word to the final prediction is obtained.

In our study, the significance of each word's contribution to the final prediction was visually represented through a text plot.
Words were highlighted in red and blue indicating that they had positive and negative influences on the prediction, respectively.
The intensity of the highlight corresponds to the degree of influence on the model's decision.

\section{Experiments}
In this section, we describe the datasets, experimental setup, and hyperparameter tuning.

\subsection{Datasets}
\label{sub:datasets}
For the evaluation of our proposed model, we utilized two publicly available code-mixed sentiment analysis datasets namely: Nepali-English \cite{pahari-shimada-2023-language} \footnote{Downloaded from \url{https://github.com/nirajpahari/nepali-english-cs-sentiment}} and Hindi-English \cite{patwa-etal-2020-semeval} \footnote{Downloaded from \url{https://ritual-uh.github.io/sentimix2020/}} code-mixed sentiment analysis dataset.
Although the official script of both Nepali and Hindi language is Devanagari, the datasets contain texts written only in romanized form.
The Nepali-English dataset contains the comments from Youtube with annotations for the sentiment analysis task.
The Hindi-English dataset contains the tweets in code-mixed language together with the annotations for the sentiment analysis task.
This dataset was introduced as a SemEval 2020 shared task.
The original dataset contains $14,000$ training sentences.
However, upon inspection it was found that $2,600$ sentences from the training set were also present in the validation set.
Hence, those sentences were removed from the training set \footnote{The `uid' field of each sentence in the validation set was compared with the `uid' field of all sentences in the training set.}.
The statistics of the dataset after the removal of repeated samples is shown in Table \ref{tab:data-statistics}.
Weighted F1-Score was used as an evaluation metric for both datasets.
\begin{table}[t]
\centering
\begin{tabular}{llll}
\hline
Language Pair  & Train  & Validation & Test  \\
\hline
Nepali-English & 8,561  & 1,071      & 1,071 \\
Hindi-English  & 12,480 & 3,000      & 3,000 \\
\hline
\end{tabular}
\caption{Data statistics for Nepali-English and Hindi-English datasets.}
\label{tab:data-statistics}
\end{table}

\subsection{Experimental Settings}
For the implementation of our proposed model for the experiments, MuRIL was employed as the base model.
\textit{`google/muril-base-cased'} model from the Huggingface library was utilized.
IndicXlit transliteration library \cite{madhani-etal-2023-aksharantar} was deployed for the transliteration of both Nepali and Hindi languages.
Adam optimizer \cite{kingma2014adam} with the learning rate of $2e-5$ was employed to update the parameters.
The batch size of $32$ and maximum length of $100$ were used for padding of the sentences.
During training, an early stopping strategy with a patience of $3$ on the performance of validation set was applied.

\subsection{Hyperparameter Tuning}
The hyperparameters, $\alpha$, $\beta$, $\gamma$, and $i$-th layer for representations impacted the performance of the proposed model.
Therefore, we performed grid-search for hyperparameter tuning.
Our initial experiments showed the importance of alignment, hence the candidate set of $\{0.7, 1.0\}$ was used for $\alpha$.
The candidate set of $\{0.3, 0.5, 0.7, 1.0\}$ was used for beta and gamma.
Similarly, the $i$-th layer for the representations alignment used the candidate set $\{7, 8, 9, 10, 11, 12\}$.

The performance of the proposed model was optimized for the Nepali-English dataset, when $\alpha = 1.0, \beta=0.7, \gamma=0.7$ and $i$-th layer $= 10$th layer.
Similarly, for the Hindi-English dataset, $\alpha = 1.0, \beta=0.7, \gamma=1.0$ and $i$-th layer $= 9$th layer performed the best.

\section{Results and Discussions}

\begin{table}[t]
\centering
\begin{tabular}{llll}
\hline
\# & Model                                                                        & Nep-Eng        & Hin-Eng        \\
\hline
1  & MuRIL (Romanized)                                                            & 70.00          & 71.00          \\
2  & MuRIL (Devanagari)                                                           & 70.38          & 71.20          \\
\hline
3  & Proposed Method                                                              & \textbf{75.09} & \textbf{73.46} \\
4  & \begin{tabular}[c]{@{}l@{}}Proposed Method\\ w/o Regularization\end{tabular} & 73.13          & 72.28          \\
5  & \begin{tabular}[c]{@{}l@{}}Proposed Method\\ w/o Alignment\end{tabular}      & 71.93          & 72.06 \\
\hline
\end{tabular}
\caption{Performance of the proposed method against that of the baseline methods.}
\label{tab:main-results}
\end{table}

\begin{table*}[t]
\begin{tabular}{ccp{0.75\textwidth}}
\hline
Method                   & Weighted F1 & Comments                                                                                                                                                         \\
\hline
\citet{liu-etal-2020-kk2018} & 75.0        & Adversarial learning on XLM-R and ensemble of models.                                                                                                             \\
\citet{srinivasan-2020-msr} & 72.6        & XLM-R model finetuned with two million generated code-mixed sentences and 90,000 real code-mixed sentences.                                                        \\
\citet{ghosh2023multitasking} & 71.61       & Transfer learning using external sentiment data, annotation of emotions labels in the original datasets and multitask learning with emotion and sentiment labels. \\
\hline
Proposed model           & 73.46       & Cross-language script transfer and alignment to share the script-specific knowledge.     \\                                                   \hline             
\end{tabular}
\caption{Performance of the proposed method on Hindi-English dataset against that of the existing methods.}
\label{tab:hin-eng-existing-result}
\end{table*}

\subsection{Results of Proposed Method}
Table \ref{tab:main-results} presents the outcomes of our proposed method in comparison to the baseline approaches.
The baseline model was the vanilla MuRIL model with a linear layer applied on top.
Row \#1 shows the performance of the baseline model with original romanized text.
Similarly, Row \#2 shows the performance with the Devanagari text (transliteration of the original data) employed on the baseline model.
Row \#3 shows the performance of the proposed model on both datasets, revealing superior results, outperforming the baseline by 5.09\% and 2.46\% on the Nepali-English and Hindi-English datasets, respectively.

In addition, we conducted an ablation study to assess the impact of regularization and alignment on our model.
Row \#4 demonstrates the performance of the proposed model without regularization, showcasing an improvement over the baseline but still underperforming in comparison to the model with regularization.
This underscores the positive impact of regularization in our proposed model.
Similarly, Row \#5 illustrates the model's performance without the alignment component, indicating a score better than the baseline but inferior to the complete proposed model.
This underscores the significance of alignment in enhancing the performance of the proposed model.

Given that the Hindi-English dataset served as a benchmark, Table \ref{tab:hin-eng-existing-result} provides a comparative analysis of the proposed method with existing state-of-the-art approaches on this dataset.
The table includes a brief description of the methods employed by existing models.
The proposed method outperformed all methods utilizing the original or additional data beyond original, with the exception of one method.
The current state-of-the-art model achieved a score of 75.0\%; however it is crucial to highlight that this performance was achieved through an ensemble of candidates.
The ensemble method considered the mean prediction from each candidate for the final prediction.
The highest score obtained by the best-performing single candidate was 73.49\%, only 0.03\% higher than our model's performance.
Hence, the performance of the proposed model was comparable to the single candidate performance of the state-of-the-art model.

\begin{table}[]
\centering
\begin{tabular}{llll}
\hline
\# & Model                                                                & Nep-Eng & Hin-Eng \\
\hline
1  & mBERT                                                                & 68.00   & 66.10   \\
2  & XLM-R                                                                & 65.00   & 69.48   \\
\hline
3  & \begin{tabular}[c]{@{}l@{}}Proposed Method\\ with mBERT\end{tabular} & 69.77   & 70.39   \\
4  & \begin{tabular}[c]{@{}l@{}}Proposed Method\\ with XLM-R\end{tabular} & 75.63   & 72.12 \\ \hline 
\end{tabular}
\caption{Performance of the proposed method on different BERT-base multilingual models.}
\label{tab:different-models-result}
\end{table}

\subsection{Effect on Different BERT-Based Multilingual Model}
To evaluate the robustness of our model across various BERT-based multilingual models, we conducted experiments using mBERT \cite{devlin-etal-2019-bert} and XLM-R \cite{conneau-etal-2020-unsupervised} as the base models.
Table \ref{tab:different-models-result} presents the performance of the proposed method, compared to those of the baseline methods with the aforementioned multilingual models as the base.

mBERT demonstrated an improvement of $1.77\%$ and $4.29\%$ for the Nepali-English and Hindi-English datasets, respectively.
Similarly, XLM-R exhibited an enhancement of $10.63\%$ and $2.64\%$ for the Nepali-English and Hindi-English datasets, respectively.
These improvements underscore the robustness of our model across different multilingual models when employed as the base model in our approach.

\begin{figure}[t]
    \centering
    \includegraphics[width=\columnwidth]{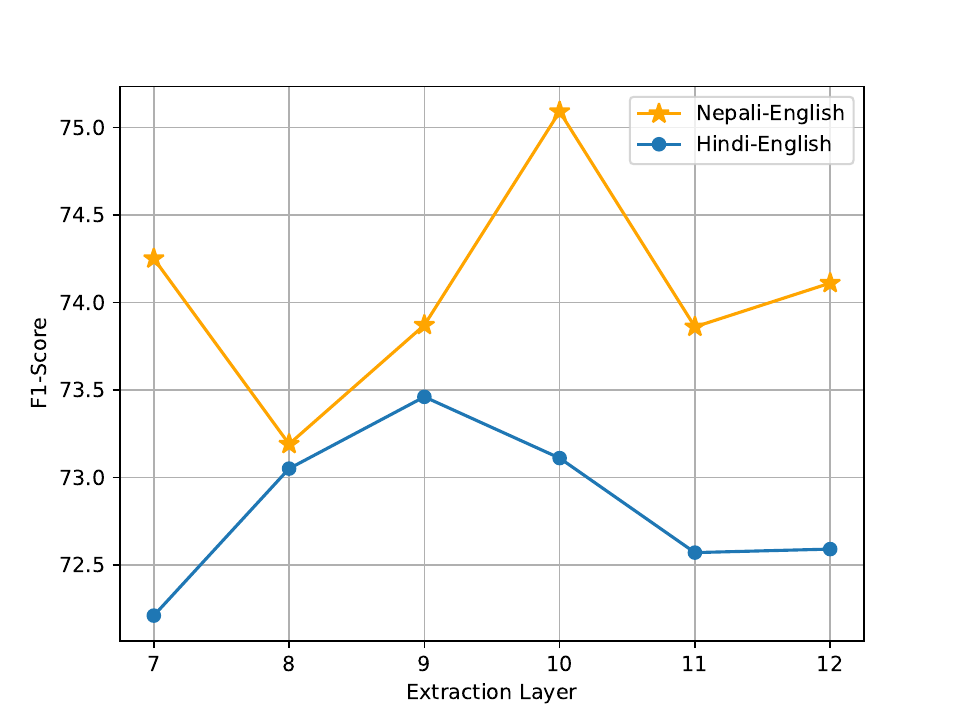}
    \caption{Performance of the proposed model with different $i$-th layer representation for alignment and regularization.}
\label{fig:ith-layer-effect}
\end{figure}

\subsection{Effect of different $i$-th layer}
To investigate the impact of utilizing representations from various layers of the language model for alignment and regularization, we maintained constant hyperparameters $\alpha$, $\beta$, and $\gamma$, while varying the $i$-th layer for both datasets.
Figure \ref{fig:ith-layer-effect} illustrates the performance of our model when representations from different layers were employed for alignment.

As explained in Section \ref{sub:cross-script-alignment}, the optimal performance was frequently achieved around the layers preceding the last layers.
This observation aligns with prior studies \cite{jalili-sabet-etal-2020-simalign, dou-neubig-2021-word, pahari2022multitask}, suggesting that the top layers may not be the most suitable for obtaining the general representations, as they might be excessively specialized for the end task.

\begin{table}[t]
\centering
\begin{tabular}{|p{\columnwidth}|}
\hline
\textbf{Sentence}:                                                           \\ sarai ramro nachnu hunchx sudhir dada best of luck                                \\
\textbf{Translation}:                                                        \\You dance really well Sudhir brother. Best of luck                                \\ \\
\hline
\textbf{True Label}: Positive                                                                          \\
\hline
\\
\textbf{Baseline Model Prediction}: Neutral\\ 
\includegraphics[width=\columnwidth]{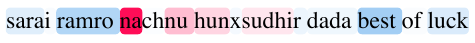}
\\
\hline
\\
\textbf{Proposed Model Prediction}: Positive\\
\includegraphics[width=\columnwidth]{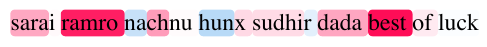}\\
\hline

\end{tabular}
\vspace{5pt} 

\begin{tabular}{|p{\columnwidth}|}
\hline
\textbf{Sentence}:                                                           \\ tapai lai pani ma chidina name fame ko lagi bolenye chudel.                                \\
\textbf{Translation}:                                                        \\I don’t know you as well. Witch who speaks only for name and fame.                                \\ \\
\hline
\textbf{True Label}: Negative                                                                          \\
\hline
\\
\textbf{Baseline Model Prediction}: Positive\\ 
\includegraphics[width=\columnwidth]{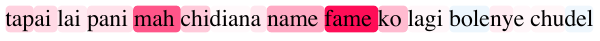}
\\
\hline
\\
\textbf{Proposed Model Prediction}: Negative\\
\includegraphics[width=\columnwidth]{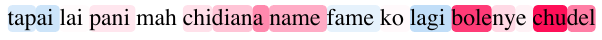}\\
\hline

\end{tabular}
\caption{Model Interpretation using SHAP. Red and blue highlight on word indicates the positive and negative influence of the word on the prediction, respectively. The intensity of highlight corresponds to the degree of influence.}
\label{tab:model-interpretation}
\end{table}

\begin{table*}[t]
\centering
\begin{tabular}{lllll}
\hline
\# & Sentence                                                                                                                          & Gold     & Baseline          & Proposed Model    \\
\hline
1  & \begin{tabular}[c]{@{}l@{}}\textbf{Original}: Ali picture quality \underline{vayan}\\ \textbf{English Translation}: The picture quality is not good enough.\end{tabular}         & Negative & Neutral           & Neutral           \\
    \cline{2-5}
   & \textbf{Modified}: Ali picture quality \underline{bhayena}                                                                                             & Negative & Neutral           & \textbf{Negative} \\
   \hline
2  & \begin{tabular}[c]{@{}l@{}}\textbf{Original}: Dherai \underline{paxi} good news \underline{sun na} paiyo\\ \textbf{English Translation}: Got to hear good news after long time.\end{tabular} & Positive & Neutral           & Negative          \\
\cline{2-5}
   & \textbf{Modified}: Dherai \underline{pachi} good news \underline{sunna} paiyo                                                                                      & Positive & \textbf{Positive} & \textbf{Positive} \\
   \hline
3  & \begin{tabular}[c]{@{}l@{}}\textbf{Original}: Game \underline{tt} babaal nai \underline{rax} ni hahaha \underline{nc}\\ \textbf{English Translation}: the game is amazing hahahaha nice\end{tabular}     & Positive & Neutral           & Negative          \\
\cline{2-5}
   & \textbf{Modified}: Game \underline{ta} babaal nai \underline{rahecha} ni hahaha \underline{nice}                                                                               & Positive & Neutral           & \textbf{Positive}\\
   \hline
\end{tabular}
\caption{Error analysis of random examples that were incorrectly predicted by the proposed model.}
\label{tab:error-analysis}
\end{table*}

\subsection{Interpretation of the Proposed Method}
The primary objective of the proposed method was to leverage the language script-specific knowledge learned during pre-training and share the knowledge across scripts for better understanding of sentences.
With random examples from the test set, we studied the cross-language script transfer and behaviour of the proposed model, as shown in Table \ref{tab:model-interpretation}. 

The first example illustrates a positive sentence from the test set.
The baseline model erroneously predicted it as a neutral sentence.
As seen from the text plot, words such as \textit{ramro (translation: good), best,} and \textit{luck} negatively influenced the prediction, whereas the subword \textit{na} from the word \textit{nachnu (translation: to dance)} positively impacted the prediction.
In Nepali, the subword \textit{na} typically acts as a prefix, changing a word's meaning to negative when added to it (e.g., \textit{ramro (translation: good)} and \textit{\textbf{na}ramro (translation: bad)}).
However, in this context, \textit{na} is not a suffix, and \textit{nachnu} is a standalone word.
The baseline model misinterpreted this, resulting in an incorrect neutral prediction.
In contrast, our proposed model accurately predicted the sentence as positive.
The representations from the Devanagari script were also shared; therefore, the model was less influenced by the subword \textit{na} and more by the words \textit{ramro (translation: good)} and \textit{best}.

Similarly, the second example depicts a negative sentence from the test set.
In this example. the word \textit{chudel (translation: witch)} is a strong negative word from the Nepali language, written in the Latin/Roman script.
The baseline model incorrectly predicted the sentence as positive, with most words positively influencing the prediction.
The word \textit{chudel} had minimal impact because the model barely comprehended its meaning.
Conversely, the proposed model correctly predicted the sentence as negative.
The plot revealed that the word \textit{chudel} significantly influenced the negative prediction.
Thus, the method effectively shared the pre-trained knowledge from the independent language script, hence capturing the semantic meaning.

\subsection{Error Analysis}
To identify the limitations of our proposed method, we analyzed samples that were misclassified by our model.
This investigation revealed that misclassification were mainly attributable to the absence of a standardized approach for writing Nepali language using the Latin/Roman script and a tendency for people to be less strict about spellings in social media settings.
Table \ref{tab:error-analysis} presents a few cases from our test set.

In the first example, the word \textit{vayan (translation: not to be)} is written incorrectly.
By `written incorrectly', we mean that this word cannot be transliterated to a correct Devanagari word.
Consequently, the prediction was neutral, whereas the actual sentiment was negative.
Modifying the word to a more correct form (i.e., \textit{bhayena (translation: not to be)}) resulted in the correct prediction.

Similarly, in the second example, the underlined words were not correct or closer to correct when transliterated using the library.
Here, we observed the unnecessary use of space between the words \textit{sun na (translation: to listen)}, which confused the model.
The correct form should have been \textit{sunna (translation: to listen)}. 
Both the baseline and proposed models made correct predictions after the underlined words were modified.

The third example highlights the casual environment of social media, where the abbreviated form \textit{nc} is used instead of the word \textit{nice}. Correcting the underlined words to their proper forms enabled the proposed model to make the correct prediction. 
These instances suggest that proper cleaning of social media text can potentially enhance performance.
These findings underscore the need for a more focused study on cleaning the romanized form of data and developing a more robust transliteration process.

\section{Conclusion}
This study was conducted to develop a model facilitating improved cross-script transfer of knowledge for code-mixing between languages utilizing different scripts.
We introduced a novel architecture featuring cross attention between script-specific representations and alignment of script representations.
Results obtained on the Nepali-English and Hindi-English datasets underscored the effectiveness of the proposed model.
Interpretation of the model using SHAP provided insights into how cross-script knowledge transfer contributes to the final prediction.
The error analysis of misclassified examples yielded valuable insights for future research directions.
It suggests the need for a comprehensive study on cleaning romanized forms of data and addressing the challenges posed by informal language expressions.
Additionally, the study emphasizes the importance of developing a more robust transliteration model that can consider the contextual nuances of the text, which enhances the transliteration process.
These insights pave the way for further refinements and advancements in the domain of cross-script transfer for code-mixing.

\bibliography{anthology,custom}

\end{document}